\documentclass[runningheads]{llncs}

\usepackage{eccv}

\usepackage{eccvabbrv}

\usepackage{graphicx}
\usepackage{booktabs}
\usepackage{array}
\usepackage{multirow}
\usepackage{array}
\usepackage{makecell}
\usepackage{xcolor}
\usepackage{bbding}

\usepackage[accsupp]{axessibility}  

\usepackage[abs]{overpic}

\def\eg{\emph{e.g}\onedot} 
\def\ie{\emph{i.e}\onedot}

\usepackage{hyperref}

\usepackage{orcidlink}

\begin{document}

\title{InstanceControl: Controllable Complex Image Generation without Instance Labeling} 

\titlerunning{InstanceControl}

\author{Xiaoyu Liu\inst{1} \and
Huan Wang\inst{1} \and
Fan Li\inst{2} \and
Zhixin Wang\inst{2} \and
Jiaqi Xu \inst{2} \and \\
Ming Liu\inst{1}$^{(}$\Envelope$^)$ \and 
Wangmeng Zuo\inst{1}}

\authorrunning{X.~Liu et al.}

\institute{Harbin Institute of Technology, Harbin, China \\
\and  HUAWEI Noah's Ark Lab, Shenzhen, China\\
\email{\{liuxiaoyu1104, whuan271199\}@gmail.com, csmliu@outlook.com, \{lifan61,wangzhixin6,xujiaqi27\}@huawei.com, wmzuo@hit.edu.cn}}

\maketitle

\vspace{-1.5mm}
\begin{center}
\textbf{Project page:} \url{https://instancecontrol.github.io/InstanceControl/}
\end{center}
\vspace{-4mm}

\begin{figure}[h]
  \centering
  \vspace{-3mm}
  \includegraphics[width=\linewidth]{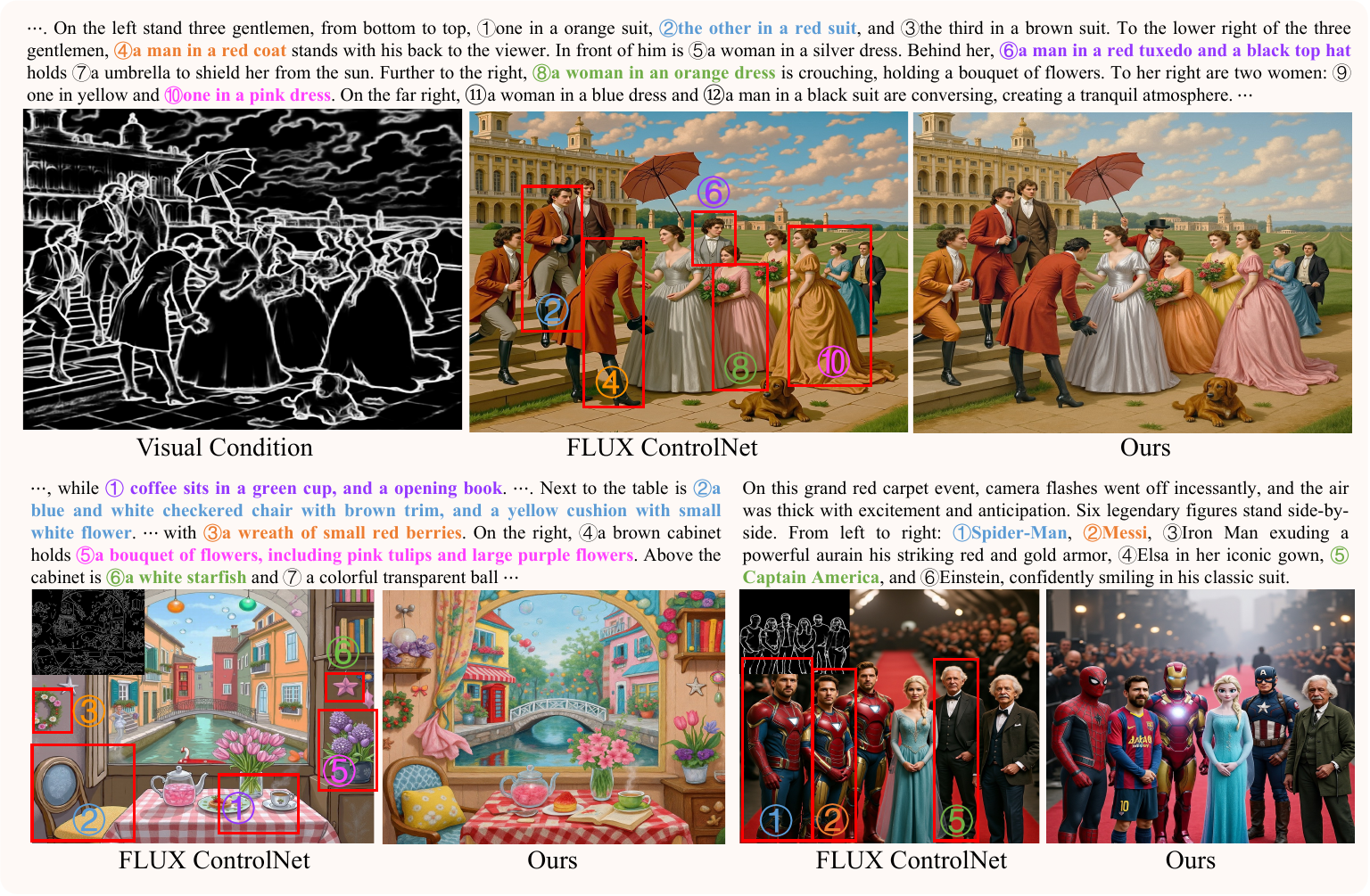}
  \vspace{-5mm}
  \caption{
Our proposed InstanceControl achieves fine-grained control over instance attributes in complex multi-instance scenarios. In contrast, FLUX ControlNet~\cite{flux2024} often struggles with attribute confusion. Incorrect instances are marked with red boxes, and the corresponding instance descriptions are also highlighted in the prompt.
  }
  \vspace{-11mm}
  \label{fig:abract_mutilinstance}
\end{figure}

\begin{abstract}

Controllable image generation methods, such as ControlNet, have demonstrated a remarkable capacity to introduce visual conditions (e.g., depth maps) to guide image generation. However, these methods often struggle with complex multi-instance scenes, frequently leading to attribute confusion among instances. While recent approaches attempt to mitigate this via manual instance labeling, such requirements are labor-intensive.
In this paper, we propose InstanceControl, a novel multi-instance controllable generation method that eliminates the need for instance labeling. We identify the primary bottleneck in existing methods as the inability to accurately associate instance descriptions with their corresponding regions within visual conditions.
To address this, we leverage the Vision-Language Model (VLM) to establish instance-level correspondences between text prompts and visual conditions. Specifically, the VLM automatically parses instance descriptions from the text prompts and simultaneously predicts instance masks based on the visual conditions. Furthermore, since the predicted masks may contain noise, we introduce an adaptive mask refinement strategy that dynamically refines these instance masks during the generation process. Extensive experiments demonstrate that our approach outperforms state-of-the-art methods, achieving superior fidelity and precise instance-level control.

\end{abstract}

\section{Introduction}
\label{sec:intro}


In recent years, controllable image generation methods~\cite{zhang2023adding, zavadski2023controlnet, bhat2023loosecontrol, peng2024controlnext, liu2024smartcontrol,rombach2022high, zhou2025dreamrenderer} like ControlNet~\cite{zhang2023adding} have gained significant traction. By introducing visual conditions (\eg, edges and depth) over text-to-image models~\cite{rombach2022high,podellsdxl,wu2025qwen}, these methods allow users to customize the spatial controls of generated images, achieving remarkable success across diverse applications (\eg, human animation~\cite{hu2024animate,zhu2024champ}, sketch-to-design~\cite{koley2024s,lin2025inkspire}, and inpainting~\cite{shin2025large}).
Building on ControlNet~\cite{zhang2023adding}, subsequent research has introduced various improvements and extensions, including efficient architectures~\cite{zavadski2023controlnet,peng2024controlnext,fang2025flexcontrol}, multi-condition fusion~\cite{hu2023cocktail,qin2023unicontrol}, and flexible control~\cite{bhat2023loosecontrol,liu2024smartcontrol}.

However, existing controllable methods are effective only in simple scenes with a limited number of instances.
As the number of instances increases, they struggle to maintain ideal generation, often leading to attribute confusion across different instances. 
To alleviate this, some recent studies~\cite{zhou2025dreamrenderer,choi2025finecontrolnet,li2025seg2any,zhang2025creatidesign,zhang2025eligen} have introduced manual instance labeling (\eg, instance descriptions and their corresponding masks or bounding boxes) to enhance fine-grained controllability in multi-instance scenarios. 
Nevertheless, these methods rely heavily on additional human annotation during inference, which is both time-consuming and labor-intensive, limiting their practicality in real-world applications.

In this work, we aim to achieve fine-grained generation in complex multi-instance scenes without requiring instance labeling during inference. 
We argue that the primary reason for the failure of existing methods lies in the inability to accurately associate each instance in the text prompt with its corresponding spatial region in the visual condition.
In contrast, humans intuitively establish such associations upon perceiving both text and visual conditions. Fortunately, recent Vision-Language Models (VLMs)~\cite{liu2023llava, Qwen2-VL, Qwen2.5-VL, wu2024deepseek,chen2024internvl,zhu2025internvl3} have demonstrated exceptional cross-modal reasoning and global understanding capabilities~\cite{yang2025thinking,yang2025r1,gou2024navigating,lin2025showui,lai2024lisa,yuan2025sa2va}, offering a potential avenue to emulate this human ability.

%

Based on this insight, we propose InstanceControl, a controllable multi-instance image generation method without the need for instance labeling. By leveraging the comprehension and reasoning capabilities of VLMs, our method automatically establishes instance-level correspondences between textual prompts and visual conditions, enabling fine-grained control in multi-instance scenarios. 

Our framework consists of two core stages: instance-level text-visual condition association and instance-aware controllable generation.
\textbf{In the first stage}, we aim to establish correspondences between text and visual conditions at the instance level, \ie, parse instance descriptions from text and predict the instance masks in the visual conditions. For predicting instance mask,  existing VLM-based grounding or referring segmentation models (\eg, LISA~\cite{lai2024lisa}) are primarily designed for RGB images, and they require strict semantic consistency between the input text and visual content.
In our task, users may employ diverse text prompts for similar visual conditions, such as using captions like \texttt{"Spider-Man, Messi, Iron Man"} to describe human edge maps (see lower right corner in \cref{fig:abract_mutilinstance}).
%
In such cases, existing referring segmentation methods fail due to the ambiguity between text and visual conditions. We take this into consideration to elaborate a tailored dataset for finetuning VLMs, enabling more robust and intelligent grounding while establishing correspondences. Furthermore, we propose a shared SEG token strategy to address the situation where an instance appears multiple times in the text.
%
%
\textbf{In the second stage}, we inject the instance-level correspondences obtained from the first stage into the generation process by binding image tokens of each instance with corresponding text tokens. 
%
%
To mitigate the adverse effects of inaccurate or noisy masks produced in the first stage, we further introduce a mask refinement module.
This module jointly leverages the confidence scores of the predicted masks and attention information of the generative model to refine the instance-level correspondences.

%

%

As shown in \cref{fig:abract_mutilinstance}, our InstanceControl demonstrates the superior ability in fine-grained control across multiple instances.
Extensive experiments are conducted on various visual conditions (\ie, canny edges, depth maps, HED~\cite{xie2015holistically}), and the results show that InstanceControl performs favorably against state-of-the-art methods in complex multi-instance scenarios.

Our contributions are listed as follows:
\begin{itemize}
\item We introduce InstanceControl, a multi-instance controllable generation method without instance labeling. It automatically associates text prompts with visual conditions at the instance level, enabling high-fidelity generation.
\item We utilize vision-language models to automatically parse instance descriptions from text prompts, while simultaneously predicting the instance masks in the visual conditions, establishing instance-level correspondences.
\item To mitigate the noise or inaccuracies in predicted masks, we introduce a mask refinement strategy to refine instance-level correspondences, ensuring more accurate multi-instance generation.


\item Experiments on multiple visual conditions demonstrate that our method outperforms existing state-of-the-art models, showcasing its effectiveness in controllable complex image generation.
\end{itemize}

\section{Related Work}

\subsection{Controllable Text-to-Image Generation}

To achieve controllable image generation, ControlNet~\cite{zhang2023adding} introduces visual conditions (\eg, depth maps) to guide image layout. Building on this, various enhancements have been proposed~\cite{hu2023cocktail, qin2023unicontrol, zavadski2023controlnet, bhat2023loosecontrol, peng2024controlnext, liu2024smartcontrol}. Specifically, UniControl~\cite{qin2023unicontrol} unifies diverse tasks via a mixture-of-experts adapter, ControlNeXt~\cite{peng2024controlnext} employs a lightweight architecture, and SmartControl~\cite{liu2024smartcontrol} adjusts control intensity to handle rough conditions. Recent efforts like OmniControl~\cite{tan2025ominicontrol} and FLUX ControlNet~\cite{flux2024} extend these capabilities to Diffusion Transformer architectures.

Despite their success, these methods struggle with attribute confusion in multi-instance scenes. To mitigate this, recent works leverage instance-level annotations for fine-grained control. Specifically, these approaches provide instance descriptions associated with instance regions such as bounding boxes or segmentation masks.
For example, FineControlNet~\cite{choi2025finecontrolnet} processes each instance description and human pose in parallel and merges them in the latent space. It also briefly explores leveraging large language models to establish instance-level correspondences for human pose.
DreamRenderer~\cite{zhou2025dreamrenderer} introduces a training-free attention mask mechanism to enforce hard binding between regions and prompts, complemented by soft binding at specific timesteps.
Meanwhile, several works integrate instance labeling into text-to-image generation. EliGen~\cite{zhang2025eligen} utilizes attention masks to regulate text-image interactions while fine-tuning via LoRA. CreatiLayout~\cite{zhang2025creatilayout} incorporates a siamese branch to inject layout guidance into the MM-DiT framework. Beyond box-level guidance, Seg2Any~\cite{li2025seg2any} incorporates pixel-wise instance labeling via attribute-isolation masks and contour extraction.
ConsistCompose~\cite{shi2025consistcompose} leverages a unified multimodal framework to address this task.
OverLayBench~\cite{li2025overlaybench} introduces a benchmark for complex spatial layouts with dense object overlaps.
Unlike these multi-instance methods, our InstanceControl achieves fine-grained control without the need for instance labeling.

\subsection{Image Segmentation with Vision-Language Models}
Vision-Language Models (VLMs)~\cite{liu2023llava,Qwen-VL,Qwen2.5-VL,Qwen2-VL,chen2024internvl,zhu2025internvl3,chen2024expanding} have shown strong capabilities in cross-modal understanding, enabling holistic reasoning over text and images. Leveraging VLMs for image segmentation~\cite{lai2024lisa,xia2024gsva,wei2024lasagna,ren2024pixellm,liu2025seg,zhang2024psalm,wei2024hyperseg,zhang2024omg,rasheed2024glamm,yuan2025sa2va} has therefore attracted growing attention. Early VLM-based grounding models~\cite{peng2023kosmos,you2023ferret,pi2023detgpt,liu2024grounding} achieve strong performance in bounding-box localization but fall short of providing the pixel-level precision required for complex visual scenes.

To address this limitation, the landmark work LISA~\cite{lai2024lisa} introduced the concept of Reasoning Segmentation, enabling precise segmentation masks from complex and implicit queries by incorporating a specialized <SEG> token and utilizing SAM~\cite{kirillov2023segment} as a mask decoder. 
GSVA~\cite{xia2024gsva} and LaSagnA~\cite{wei2024lasagna} enhanced model robustness for processing generalized and complex language instructions.
To refine mask quality and spatial sensitivity, LISA++\cite{yang2023lisa++} and LIRA\cite{li2025lira} introduced improved training strategies and local interleaved region assistance. 
PixelLM~\cite{ren2024pixellm} and HiMTok~\cite{wang2025himtok} proposed multi-scale reasoning and hierarchical mask tokens to effectively handle multiple targets.
Seg-Zero~\cite{liu2025seg} and Seg-R1~\cite{you2025seg} employed reinforcement learning to guide multi-step reasoning, thereby reducing the reliance on expensive annotations.
Meanwhile, PSALM~\cite{zhang2024psalm} and HyperSeg~\cite{wei2024hyperseg} unified diverse segmentation tasks into a generative MLLM paradigm. 
Building upon reasoning segmentation, recent research has further expanded the functional boundaries of pixel-level understanding. 
GLaMM~\cite{rasheed2024glamm} and Omg-llava~\cite{zhang2024omg} bridged captioning and grounding, enabling models to describe and segment objects simultaneously. 
Sa2VA~\cite{yuan2025sa2va} provides a unified framework for dense image and video grounding. When integrated with advanced VLMs, it delivers competitive performance in referring segmentation.
%
%
Existing grounding approaches focus on RGB images with strict text-visual alignment. We aim to develop an analogous grounding mechanism for visual conditions.

\section{Methodology}

Controllable image generation aims to achieve spatial control over generated images based on both the text prompt $\mathbf{p}$ and the visual condition $\mathbf{c}$.
However, in multi-instance scenarios, previous approaches often struggle to maintain fine-grained attribute control for each instance, frequently leading to attribute leakage across different instances. We attribute this limitation to the lack of instance-level correspondences between text prompts and visual conditions.

\begin{figure}[tb]
  \centering
  \includegraphics[width=1\linewidth]{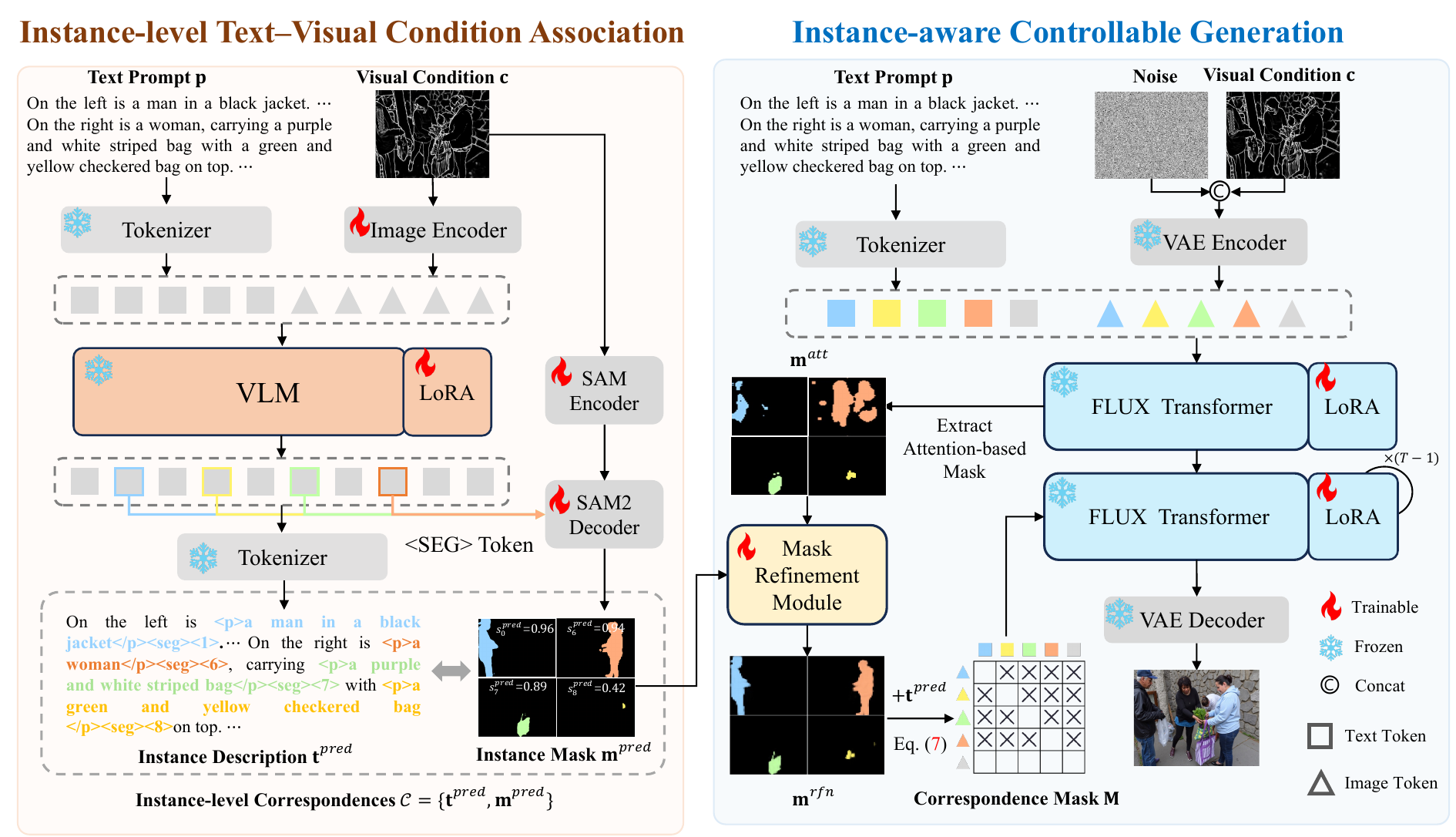}
  \caption{ The proposed InstanceControl framework operates in two stages: instance-level text-visual condition association and instance-aware controllable generation. In the first stage, a VLM is used to establish instance-level correspondences $\mathcal{C}$ between text prompts and visual conditions. To mitigate noise in predicted masks, the second stage introduces a mask refinement module that adjusts $\mathbf{m}_{i}^\mathit{pred}$ to $\mathbf{m}_{i}^\mathit{rfn}$ based on confidence scores and attention-based masks. These refined correspondences are incorporated into the generation process using a correspondence mask. }
  \vspace{-5mm}
  \label{fig:method_mutilinstance}
\end{figure}

To address this, we propose a two-stage framework, as illustrated in \cref{fig:method_mutilinstance}, comprising instance-level text-visual condition association and instance-aware controllable generation. In the first stage, we utilize the vision-language model to establish instance-level correspondences between text prompts and visual conditions. The instance-level correspondences contain information for $N$ entities, where each entity consists of two parts: the instance description $\mathbf{t}$ (extracted from the text prompt $\mathbf{p}$) and the instance mask $\mathbf{m}$, denoted as:
\begin{equation}
\mathcal{C} = [(\mathbf{t}_1,\mathbf{m}_1),(\mathbf{t}_2,\mathbf{m}_2), \ldots, (\mathbf{t}_N,\mathbf{m}_N)].
\end{equation}
In the second stage, we inject the learned instance-level correspondences into the generation process, enabling more accurate and finer-grained image generation.

\subsection{Instance-level Text-Visual Condition Association }
\label{sec:mutilinstance_association}
To obtain the instance-level correspondence $\mathcal{C}$ between the visual condition $\mathbf{c}$ and the text prompt $\mathbf{p}$, we need to parse the instance descriptions in the text and simultaneously predict their corresponding masks in the visual condition. 
A key challenge arises with complex prompts: descriptions of different instances are often interleaved with verbs or prepositions, while the descriptions of the same instance may appear across the sentence. To address this, we extract all descriptive noun phrases for each instance at different positions in the prompt, and link them with a unique instance ID. 
The overall pipeline of the first stage is related to the multi-object referring segmentation.
Accordingly, we adopt the Sa2VA~\cite{yuan2025sa2va} architecture, which integrates SAM~\cite{kirillov2023segment} with VLM, as our backbone.

\noindent\textbf{Instance Description Parsing}.
Given the visual condition $\mathbf{c}$ and text prompt $\mathbf{p}$ , the VLM $\mathcal{F}$ generates a text response $\mathbf{y}_\mathit{txt}^\mathit{pred}$:
\begin{equation}
\mathbf{y}_\mathit{txt}^\mathit{pred}=\mathcal{F}(\mathbf{c}, \mathbf{p}),
\end{equation}
where, the response $\mathbf{y}_\mathit{txt}^\mathit{pred}$ follows a fixed format, for example:
``\texttt{On the left is <p>a man in a black jacket</p><seg><1>.}''.
Here, \texttt{<p>} and \texttt{</p>} denote the beginning and end of each phrase, and each phrase is followed by a \texttt{<SEG>} token and an instance identity \texttt{<id>}. 
The predicted instance description $\mathbf{t}_{i}^\mathit{pred}$ is then constructed by aggregating all phrases associated with the instance identity \texttt{<i>}.

\noindent\textbf{Shared SEG Token}.
Each $\texttt{<SEG>}$ token provides a coarse localization cue for its corresponding phrase predicted by the VLM. We project the final-layer hidden embeddings of these tokens via an MLP to generate query representations $\mathbf{h}$.
When an instance is described by multiple phrases, the model generates several $\texttt{<SEG>}$ tokens for a single entity. To ensure consistent mask predictions, we adopt a Shared SEG Token (SST) strategy. Specifically, $\texttt{<SEG>}$ tokens sharing the same instance identity are grouped, and their query representations are aggregated into a unified representation $\mathbf{h}_{i}$, \ie,
\begin{equation}
\mathbf{h}_{i} =   \text{Concat}({\mathbf{h}}_{i}^1, {\mathbf{h}}_{i}^2, \dots, {\mathbf{h}}_{i}^\mathit{M}) ,
\end{equation}
where $M$ denotes the number of \texttt{<SEG>} tokens corresponding to instance $i$.

\noindent\textbf{Instance Mask Decoding}.
We leverage SAM to refine these unified representations into dense instance masks. Specifically, the SAM image encoder $\mathcal{F}_\mathit{enc}$ extracts dense image features $\mathbf{f}$ from the visual condition. We then input the aggregated instance query $\mathbf{h}_i$ and the visual features $\mathbf{f}$ into the SAM decoder $\mathcal{F}_{\text{dec}}$ to predict the final instance mask $\mathbf{m}_i^\mathit{pred}$ and confidence score $s_i^\mathit{pred}$:
\begin{equation}
\mathbf{f} = \mathcal{F}_\mathit{enc}(\mathbf{c}), \quad (\mathbf{m}_{i}^\mathit{pred}, {s}_i^\mathit{pred})= \mathcal{F}_{dec}(\mathbf{h}_{i}, \mathbf{f}).
\end{equation}

\noindent\textbf{Training Objectives}.
The total loss function $\mathcal{L}$ is a combination of the text generation loss $\mathcal{L}_{txt}$ and the mask prediction loss $\mathcal{L}_{mask}$:
\begin{equation}
\begin{gathered}
\mathcal{L} = \mathcal{L}_{txt} + \mathcal{L}_{mask}, \quad \mathcal{L}_{txt} = \text{CE}(\mathbf{y}_\mathit{txt}^\mathit{pred}, \mathbf{y}_{txt}^{gt}), \\
\mathcal{L}_{mask} = \sum\limits_{i=0}^{N} \left( \lambda_{bce} \text{BCE}(\mathbf{m}_{i}^\mathit{pred}, \mathbf{m}_{i}^{gt}) + \lambda_{dice} \text{DICE}(\mathbf{m}_{i}^{pred}, \mathbf{m}_{i}^{gt}) \right),
\end{gathered}
\end{equation}
where ${\text{CE}}(\cdot)$ denotes the autoregressive cross-entropy loss for text generation, while ${\text{BCE}}(\cdot)$ and ${\text{DICE}}(\cdot)$ represent the pixel-wise binary cross-entropy loss and DICE loss, respectively. The terms $\lambda_{\text{bce}}$ and $\lambda_{\text{dice}}$ are the loss weights.

\subsection{Instance-aware Controllable Generation. }

In this section, we integrate the learned instance-level correspondences $\mathcal{C} = \{(\mathbf{t}_{i}^\mathit{pred}, \mathbf{m}_{i}^\mathit{pred})\}_{i=1}^N$ into the generation process.
Unlike prior multi-instance methods that rely on precise, user-provided masks, our approach utilizes predicted masks $\mathbf{m}_{i}^\mathit{pred}$. These masks maybe contain noise, such as incomplete regions or localization offsets.
Directly imposing these masks as hard constraints often yields sub-optimal results. This motivates us to adaptively utilize the predicted masks: the model should strictly adhere to $\mathbf{m}_{i}^\mathit{pred}$ when accurate, but relax the constraint when it is unreliable to avoid over-fitting to erroneous regions.

\noindent\textbf{Mask Refinement Module}. To implement this, we exploit three key signals. First, the confidence score $s_i^\mathit{pred}$ serves as an indicator of mask reliability. Second, we extract an attention-based mask $\mathbf{m}_i^{\mathit{att}}$ by averaging cross-attention maps associated with $\mathbf{t}_{i}^\mathit{pred}$ across selected layers, capturing the intrinsic spatial intent of the diffusion model. Third, we observe a complementary relationship between $\mathbf{m}_{i}^\mathit{pred}$ and $\mathbf{m}_{i}^\mathit{att}$.
Motivated by these observations, we employ a mask refinement module $f$ to integrate these signals. At each denoising timestep $t$, this module takes the predicted mask $\mathbf{m}^{\mathit{pred}}_{i}$, the attention-based mask $\mathbf{m}^{\mathit{att}}_{i}$, and the confidence score $s_i^\mathit{pred}$, conditioned on the image latent features $\mathbf{H}_{\mathit{img}}$, to produce a rectified mask $\mathbf{m}_{i}^{\mathit{rfn}}$:
\begin{equation} 
\mathbf{m}_{i}^\mathit{rfn} = \mathit{f} (\mathbf{m}^{\mathit{pred}}_{i}, \mathbf{m}^{\mathit{att}}_{i}, s_i^\mathit{pred}, \mathbf{H}_{\mathit{img}}), \end{equation}
where the mask refinement module is built upon a lightweight U-Net framework, with each block incorporating a ResNet module along with both intra-instance and inter-instance attention mechanisms.
As shown in \cref{fig:method_mutilinstance}, the refined mask $\mathbf{m}_{i}^\mathit{rfn}$ maintains consistency with $\mathbf{m}^{\mathit{pred}}_i$ when $s_i^\mathit{pred}$ is high. Conversely, when $s_i^\mathit{pred}$ is low, the refined mask $\mathbf{m}_{i}^\mathit{rfn}$ aligns more closely with the attention-based mask $\mathbf{m}^{\mathit{att}}_{i}$.
This indicates that the refined mask $\mathbf{m}^{\mathit{rfn}}_{i}$ is plausible, and we utilize $\mathbf{m}_{i}^\mathit{rfn}$ to replace the predicted mask $\mathbf{m}^{\mathit{pred}}_{i}$.

\noindent\textbf{Correspondences Mask Construction}.
To seamlessly incorporate these refined correspondences $\{(\mathbf{t}_{i}^\mathit{pred}, \mathbf{m}^{\mathit{rfn}}_{i})\}_{i=1}^N$ into the generation process, we follow previous multi-instance methods\cite{li2025seg2any,zhang2025eligen} to construct a correspondence mask $\mathbf{M}$ that constrains the image-text attention. This mask $\mathbf{M}$ ensures that image tokens of specific instances attend exclusively to their corresponding text tokens, thereby mitigating inter-instance interference. Specifically, let $\mathcal{T}_i$ denote the set of text tokens derived from the instance description $\mathbf{t}_{i}^\mathit{pred}$, while $\mathcal{I}_i$ and $\mathcal{I}_\mathit{bg}$ represent image token sets for the refined instance mask $\mathbf{m}^{\mathit{rfn}}_{i}$ and background regions, respectively. The correspondences mask $\mathbf{M} \in \mathbb{R}^{L_{\mathit{img}} \times L_{\mathit{txt}}}$, where $L_{\mathit{img}}$ and $L_{\mathit{txt}}$ denote the number of image and text tokens, is defined as:

\begin{equation}
\mathbf{M}[q,k] =
\begin{cases}
\mathbf{m}^{\mathit{rfn}}_{i}, & \text{if } q \in \mathcal{I}_i \text{ and } k \in \mathcal{T}_i , i \in [1, N] \\
1, & \text{if } q \in \mathcal{I}_\mathit{bg} \\
0, & \text{otherwise}
\end{cases}.
\end{equation}
By integrating this correspondence mask into the attention mechanism, the modified calculation is formulated as:
\begin{equation}\text{Attention}(\mathbf{Q}, \mathbf{K}, \mathbf{V}) = \text{Softmax} \left( \frac{\mathbf{Q}\mathbf{K}^T}{\sqrt{d}} + \log(\mathbf{M}) \right) \mathbf{V},
\end{equation}
where $\mathbf{Q}$, $\mathbf{K}$, and $\mathbf{V}$ represent the query and key projections. When $\mathbf{M}[q,k] \to 1$, $\log(\mathbf{M})$ approaches $0$, recovering to vanilla attention and preserving token interactions. Conversely, when $\mathbf{M}[q,k] \to 0$, $\log(\mathbf{M})$ approaches $-\infty$, effectively suppressing irrelevant information by nullifying the corresponding attention weights.

\noindent\textbf{Training Objectives}.
We employ the rectified flow-matching loss and the mask loss to train our mask refinement module. The total objective $\mathcal{L}$ is a combination of the flux loss $\mathcal{L}_{flux}$ and the mask prediction loss $\mathcal{L}_{mask}$:
\begin{equation}
\begin{gathered}
 \mathcal{L}_{flux} = \mathbb{E}_{ \mathbf{t},\mathbf{z}_0, \mathbf{p}, \mathbf{c}, \epsilon \sim \mathcal{N}(0,1)}[\| v_\theta(\mathbf{z}_\mathit{t}, \mathbf{t}, \mathbf{p}, \mathbf{c})-(\varepsilon-\mathbf{z}_0) \|_2^2],
    \label{eqn:SmartControl_ldmloss} \\
\mathcal{L}_{mask} = \sum\limits_{i=0}^{N} \left( \lambda_{bce} \text{BCE}(\mathbf{m}^{\mathit{rfn}}_{i}, \mathbf{m}_{i}^{gt}) + \lambda_{dice} \text{DICE}(\mathbf{m}^{\mathit{rfn}}_{i}, \mathbf{m}_{i}^{gt}) \right),
\end{gathered}
\end{equation}                   
where $v_\theta$ denotes our model and $\mathbf{z}_0$ represents the latent embedding of the real image. $\mathbf{z}_t$ is a linear interpolation between $\mathbf{z}_0$ and noise $\varepsilon \sim \mathcal{N}(0, 1)$. Here, $t \in [0, 1]$ denotes the timestep of the flow-matching process.

\subsection{Data Construction}
Existing datasets suffer from two major limitations: either the prompts are overly brief with too few instances~\cite{rasheed2024glamm}, or the instance descriptions are excluded from the concise text prompts~\cite{li2025seg2any,zhang2025creatilayout}. To address these limitations, we adopt the following strategies to construct our training data.
(1) Image and Mask Selection. We curate 50K images and their corresponding masks from SAM~\cite{kirillov2023segment}, COCO~\cite{lin2014microsoft}, and UniWorld-V1~\cite{lin2025uniworld}. We filter out invalid or irrelevant masks to maintain an instance density of 0--40 per image (averaging 11.64) and extract corresponding canny edges, depth maps, and HED~\cite{xie2015holistically}.
(2) Detailed Long Prompt. We employ Gemini 2.5 Pro to generate comprehensive descriptions (averaging 183.15 tokens per prompt), ensuring a precise characterization of every instance.
(3) Correspondence Generation. To build accurate instance-level correspondences, we apply Gemini 2.5 Pro to extract relevant nouns for each object from the long prompts, guided by the instance masks overlaid on the RGB images.
(4) Robust Data Augmentation. To improve robust and intelligent grounding, we construct over 1,000 analogy-based samples. We use Nano Banana~\cite{nano} to edit instances into semantically similar categories, creating the unaligned pairs.
The details of dataset construction are provided in the supplementary material.

\section{Experiments}

\subsection{Experimental Details}

\noindent\textbf{Baselines.}
We compare our method with state-of-the-art controllable text-to-image frameworks, including FLUX ControlNet~\cite{flux2024} and DreamRenderer~\cite{zhou2025dreamrenderer}, across various visual conditions such as canny edges, depth maps, and HED maps. Furthermore, we evaluate our approach against several multi-instance generation methods, specifically EliGen~\cite{zhang2025eligen}, CreatiLayout~\cite{zhang2025creatidesign}, and Seg2Any~\cite{li2025seg2any}. While EliGen and CreatiLayout rely on box-text correspondences, Seg2Any operates on segmentation-text correspondences. Notably, among all the baselines, only FLUX ControlNet~\cite{flux2024} does not require instance-level labeling.

\noindent\textbf{Evaluation Metrics.}
Following prior works~\cite{li2025seg2any,zhang2025creatidesign}, we assess the quality of generated images across three dimensions: Mean Intersection over Union (MIoU), Region-wise Quality, and Global-wise Quality. Specifically, \textit{MIoU} is employed to assess the spatial alignment between the generated objects and the ground-truth masks. 
For Region-wise Quality, we crop instance-level regions according to the ground-truth masks. We then calculate the \textit{Local CLIP}~\cite{xie2025fg} and \textit{Accuracy}~\cite{li2025seg2any} metrics to evaluate the alignment between instance regions and their corresponding instance descriptions. The \textit{Accuracy} metric leverages Qwen2-VL-72B~\cite{Qwen2-VL} in a Visual Question Answering (VQA) framework to perform a fine-grained assessment of spatial, color, text, and shape fidelity.
Finally, we employ \textit{FID}~\cite{heusel2017gans} and \textit{ImageReward}~\cite{xu2023imagereward} to assess the global image quality. Notably, \textit{CLIP Score}~\cite{radford2021learning} and \textit{PickScore}~\cite{kirstain2023pick} are omitted due to their 77-token text length limitation.

\noindent\textbf{Evaluation Benchmarks.}
We evaluate our model on the MIG-Eval and COCO-POS benchmarks. Specifically, MIG-Eval comprises 5,400 images derived from a disjoint subset of our constructed dataset. Each sample in this benchmark is featured with a comprehensive text prompt, instance-level masks, and the corresponding instance descriptions extracted from the text prompt.

\noindent\textbf{Implementation Details.}
Our training process consists of two primary stages. In the first stage, we employ the pretrained Sa2VA~\cite{yuan2025sa2va} as our backbone. Low-Rank Adaptation (LoRA) modules, with a rank of 256, are integrated into the VLM, image encoder, and the SAM encoder and decoder. We train for 30k steps with a batch size of 64 using AdamW~\cite{loshchilov2017decoupled}. The learning rate starts at $4 \times 10^{-5}$ with a 5\% linear warmup and a weight decay of 0.05. 
Following Sa2VA~\cite{yuan2025sa2va}, the loss weights $\lambda_{\mathrm{bce}}$ and $\lambda_{\mathrm{dice}}$ are set to 2.0 and 0.5, respectively.
In the second stage, we build upon the pretrained FLUX.1-Canny~\cite{flux2024}, FLUX.1-Depth~\cite{flux2024}, and XLabs HED ControlNet. We first perform LoRA fine-tuning on the ground-truth masks by applying LoRA to all linear layers within each DiT block. This process lasts 80,000 steps with a batch size of 4, utilizing a cosine learning rate schedule that starts at $1 \times 10^{-4}$. Subsequently, the mask refinement module is trained on the predicted masks for 10,000 steps under the same learning rate configuration. All experiments are conducted on four NVIDIA A6000 GPUs.

\begin{figure}[t!]
  \centering
  \includegraphics[width=0.99\linewidth]{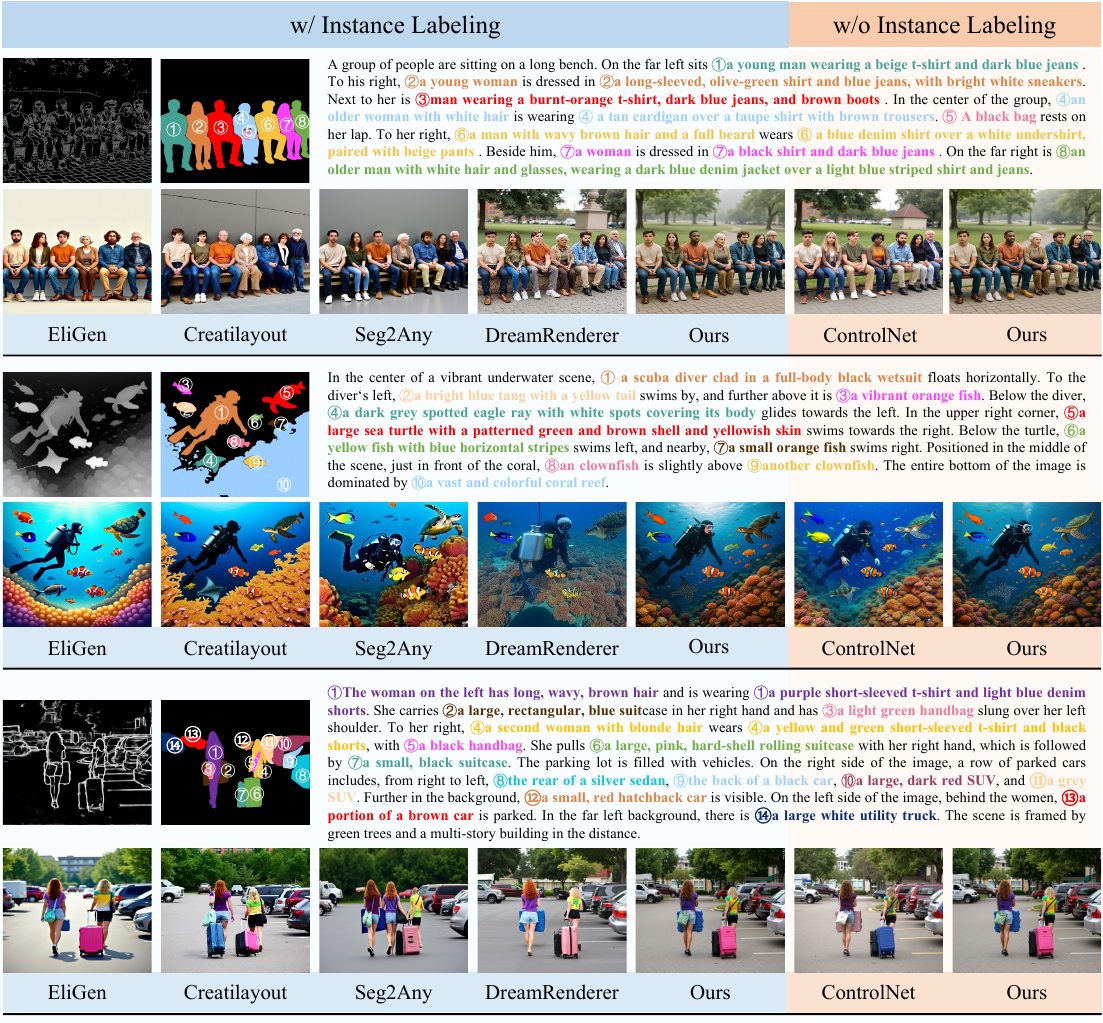}
  \vspace{-2mm}
  \caption{Qualitative comparisons in multi-instance scenarios across various image prompts and visual conditions. Our InstanceControl achieves significantly finer attribute control for each instance compared to existing methods.}
  \vspace{-1mm}
  \label{fig:com_mutilin}
\end{figure}

\begin{table*}[t!]
\centering
\caption{
Quantitative comparison of controllable text-to-image generation methods in multi-instance scenarios on MIG-Eval. The best results are highlighted with \textbf{bold}.}
\vspace{-1mm}
\resizebox{\textwidth}{!}{
\begin{tabular}{clcccccccc} 
\toprule

&\multirow{3}{*}{Methods} & \multirow{3}{*}{MIoU} &\multicolumn{5}{c}{Region-wise Quality
} &\multicolumn{2}{c}{Global-wise Quality}\\
\cmidrule(lr){4-8}
\cmidrule(lr){9-10}
  & & & \makecell[c]{Local \\ CLIP$\uparrow$}  & Spatial$\uparrow$ & Color$\uparrow$ & Shape$\uparrow$  & Texture$\uparrow$  & IR$\uparrow$  &  FID$\downarrow$   \\
\midrule
\multirow{9}{1.5cm}{\centering w/ Instance Labeling} & EliGen~\cite{zhang2025eligen} &  0.6104 & 17.49 & 91.97\% & 84.98\% & 85.59\% & 87.93\% & 0.3708 & 27.8240 \\
 & Creatilayout~\cite{zhang2025creatidesign}   &0.5247& 16.84&88.16\% & 80.87\% & 83.00\%  & 85.91\% & 0.2979&21.3988 \\
 & Seg2Any~\cite{li2025seg2any}  & 0.8316 & 18.62 &91.71\% &87.25\% &88.39\% &90.14\% &0.0893&21.2305 \\
\cmidrule(lr){2-10}
& DreamRenderer~\cite{zhou2025dreamrenderer} (Canny)
& 0.6497 & 16.25 & 80.74\% & 67.55\% & 69.20\% & 74.68\% & 0.1877 &14.5079 \\
& InstanceControl (Canny) & \textbf{0.8381} & \textbf{18.97} & \textbf{95.34\%} & \textbf{92.11\%}& \textbf{92.29\%} & \textbf{93.66\%} & \textbf{0.2940}& \textbf{9.9340}\\
\cmidrule(lr){2-10}
& DreamRenderer~\cite{zhou2025dreamrenderer} (Depth) & 0.7738  & 17.88 & 89.52\%  & 80.91\%  & 81.45\% &86.69\% & \textbf{0.3206}  & 15.3637 \\
 & InstanceControl (Depth) & \textbf{0.8212} & \textbf{18.73} & \textbf{95.06\%} & \textbf{91.35\%} & \textbf{92.08\%} & \textbf{93.47\%}  & 0.2921 & \textbf{11.9211}\\
 \cmidrule(lr){2-10}
 & DreamRenderer~\cite{zhou2025dreamrenderer} (Hed)
 & 0.7060 & 15.08 & 78.93\% &60.56\% &63.18\%& 68.79\%& 0.0392& 22.6739\\
 & InstanceControl (Hed) & \textbf{0.8504}& \textbf{19.08} & \textbf{95.77\%} & \textbf{92.58\%}& \textbf{92.77\%} & \textbf{94.10\%}  & \textbf{0.3256} & \textbf{10.2175} \\

\midrule
 \multirow{6}{1.5cm}{\centering w/o Instance Labeling} & FLUX ControlNet~\cite{flux2024} (Canny)
 & 0.6526 & 16.48& 84.67\% &73.30\% &73.93\% &79.24\%  &0.2558&14.0427\\
  & InstanceControl (Canny)
 & \textbf{0.8250} & \textbf{18.51}& \textbf{93.54\%} & \textbf{87.78\%} & \textbf{88.19\%} & \textbf{90.88\%} & \textbf{0.2847} & \textbf{10.0264} \\
  \cmidrule(lr){2-10}
 & FLUX ControlNet~\cite{flux2024} (Depth)
 & 0.7782 &17.73&90.14\% &75.87\%& 78.20\%& 82.33\%&\textbf{0.3216}&14.1077\\
 & InstanceControl (Depth)
 &  \textbf{0.8116} & \textbf{18.18} & \textbf{92.87\%} & \textbf{86.11\%} & \textbf{86.49\%} & \textbf{89.91\%} & 0.2961 & \textbf{11.9181} \\
  \cmidrule(lr){2-10}
 & FLUX ControlNet~\cite{flux2024} (Hed) &0.6817&15.67& 83.09\%& 70.15\%& 70.51\%& 77.08\%& 0.2314& 20.0755 \\
 & InstanceControl (Hed)
 &  \textbf{0.8472} & \textbf{18.74} & \textbf{94.64\%} & \textbf{88.75\%} & \textbf{89.55\%} &  \textbf{91.18\%}& \textbf{0.3268} & \textbf{10.14} \\
\bottomrule
\end{tabular}
\label{tab:comparison_mutilin_eval}}
\vspace{-4mm}
\end{table*}

\begin{figure}[t]
  \centering
  \includegraphics[width=\linewidth]{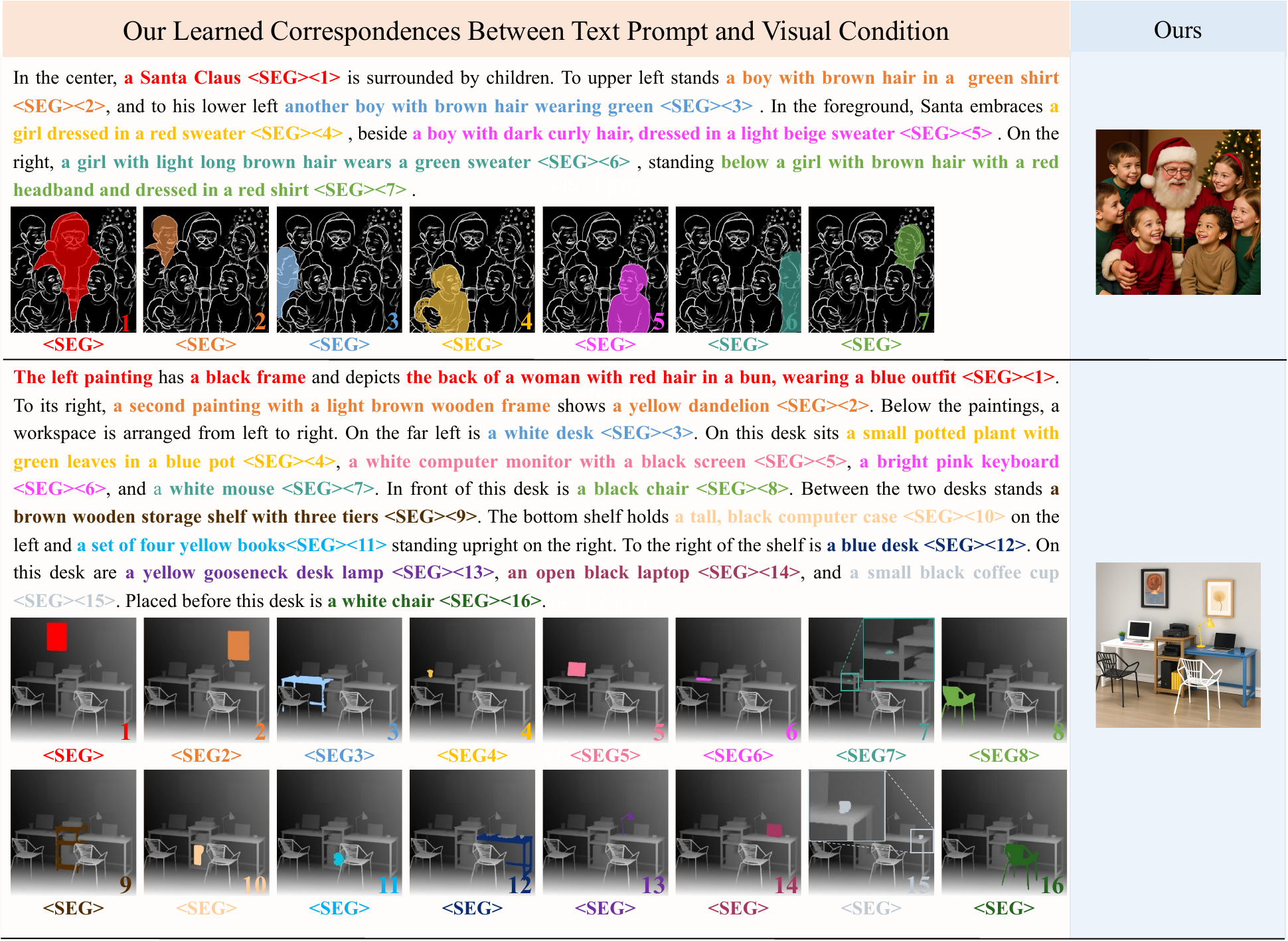}
  \vspace{-5mm}
  \caption{
Visualization of learned correspondences between text prompts and visual conditions, alongside generated images under different instance numbers.
  }
  \vspace{-2mm}
  \label{fig:com_instance_num}
\end{figure}

\subsection{Comparison with Existing Methods}

\noindent\textbf{Quantitative Comparison.}
(1) We conduct comprehensive experiments to evaluate the effectiveness of the proposed method, with quantitative results presented in \cref{tab:comparison_mutilin_eval}. The comparison is categorized into two groups: methods with and without instance labeling. To ensure a thorough evaluation, we also introduce a variant of our approach that utilizes ground-truth correspondences during the second stage.
These instance-labeled methods achieve reasonable performance by relying on precise annotations. 
Our corresponding variant consistently outperforms DreamRenderer~\cite{zhou2025dreamrenderer} by incorporating the correspondence mask through LoRA-based fine-tuning.
Conversely, existing label-free methods often fail to accurately map text to visual regions. By leveraging a VLM to bridge this gap, our approach significantly surpasses FLUX ControlNet~\cite{flux2024}. For instance, for canny conditions, our method achieves a $\sim$12.3\% gain in both \textit{Accuracy} and \textit{Local CLIP} Score. 
Remarkably, our approach even surpasses instance-labeled methods such as CreatiLayout~\cite{zhang2025creatidesign}. 
Although our method yields a slightly lower ImageReward~\cite{xu2023imagereward} than FLUX ControlNet~\cite{flux2024} under depth conditions, this is likely because the metric favors visual smoothness over realistic textures.
(2) Additionally, quantitative results on the COCO-POS benchmark in \cref{tab:comparison_mutilin_coco} show that our approach outperforms FLUX ControlNet~\cite{flux2024} across nearly all metrics, further demonstrating the effectiveness of the proposed approach on this dataset.
(3) Furthermore, we randomly select 300 images from the MIG-Eval and evaluate our method against unified understanding and generation models, including the open-source Qwen-Image ControlNet~\cite{qwenimagecontol} and the closed-source Nano Banana~\cite{nano}. Quantitative results in \cref{tab:comparisoninstance_ugm} reveal that our proposed model consistently outperforms these unified methods across all metrics.
(4) To mitigate potential in-domain bias, we provide additional quantitative results on the out-of-domain benchmark, HiCo-7K~\cite{cheng2024hico}, in the supplementary material.

\begin{table*}[t!]
\centering
\caption{
 Quantitative comparison on COCO-POS dataset.
}
\vspace{-3mm}
\resizebox{\textwidth}{!}{
\begin{tabular}{clcccccccc} 
\toprule

&\multirow{3}{*}{Methods} & \multirow{3}{*}{MIoU$\uparrow$} &\multicolumn{5}{c}{Region-wise Quality
} &\multicolumn{2}{c}{Global-wise Quality}\\
\cmidrule(lr){4-8}
\cmidrule(lr){9-10}
  & & & \makecell[c]{Local \\ CLIP$\uparrow$}  & Spatial$\uparrow$ & Color$\uparrow$ & Shape$\uparrow$  & Texture$\uparrow$  & IR$\uparrow$  &  FID$\downarrow$   \\
\midrule
\multirow{9}{1.5cm}{\centering w/ Instance Labeling} & EliGen~\cite{zhang2025eligen} &  0.5076 & 16.13 & 93.34\% & 76.59\% & 83.22\% & 82.39\% & 0.3006 & 42.5613 \\
 & Creatilayout~\cite{zhang2025creatidesign}   &0.4664& 15.84&90.74\% & 72.65\% & 76.74\%  & 75.49\% & 0.2196&32.3662 \\
 & Seg2Any~\cite{li2025seg2any}  & 0.7499 & 16.39 &89.98\% &79.36\% &82.39\% &83.66\% &-0.0347&37.4553 \\
\cmidrule(lr){2-10}
& DreamRenderer~\cite{zhou2025dreamrenderer} (Canny)
 & 0.5926 & 15.20 & 86.60\% & 62.04\% & 65.12\% & 64.23\% & 0.1367 &25.3367 \\
  & InstanceControl (Canny) & \textbf{0.7757} & \textbf{17.00}&\textbf{92.77\%} &\textbf{85.92\%}& \textbf{88.70\%} & \textbf{91.83\%} &\textbf{0.1706}&\textbf{20.8419}\\
    \cmidrule(lr){2-10}
 & DreamRenderer~\cite{zhou2025dreamrenderer} (Depth)
 & 0.7500  & 16.42 & 90.76\%  & 75.51\%  & 83.39\% &79.86\% & \textbf{0.1781}  & 26.1377 \\
  & InstanceControl (Depth) & \textbf{0.7938} & \textbf{16.97} & \textbf{92.10\%} & \textbf{83.78\%} & \textbf{88.29\%} & \textbf{88.35\%}  &0.1150 & \textbf{22.8134}\\
    \cmidrule(lr){2-10}
 & DreamRenderer~\cite{zhou2025dreamrenderer} (Hed)
 & 0.6681 & 14.77 & 84.79\% &60.41\% &67.11\%& 62.25\%& 0.0599& 29.4211\\
 & InstanceControl (Hed) & \textbf{0.8249}& \textbf{17.26} & \textbf{93.24\%} & \textbf{85.82\%}& \textbf{87.29\%} & \textbf{91.62\%}  & \textbf{0.1310} & \textbf{20.9018} \\

\midrule
 \multirow{6}{1.5cm}{\centering w/o Instance Labeling} & FLUX ControlNet~\cite{flux2024} (Canny)
 & 0.5639 & 15.09& 84.46\% &63.50\% &66.28\% &65.63\%  &0.1599&24.9675\\
  & InstanceControl (Canny)
 & \textbf{0.7688}  & \textbf{16.58}& \textbf{91.32\%} & \textbf{78.19\%} & \textbf{82.56\%} & \textbf{85.49\%} & \textbf{0.1648} & \textbf{20.8162} \\
\cmidrule(lr){2-10}
 & FLUX ControlNet~\cite{flux2024} (Depth)
 & 0.7400 &16.14&90.32\% &66.47\%& 74.75\%& 71.97\%&\textbf{0.2104}&25.6307\\
  & InstanceControl (Depth)
 &   \textbf{0.7894}& \textbf{16.50}& \textbf{90.58\%}&  \textbf{75.61\%}& \textbf{82.61\%} & \textbf{82.53\%}& 0.1170&  \textbf{22.7549 } \\
\cmidrule(lr){2-10}
 & FLUX ControlNet~\cite{flux2024} (Hed) &0.6315&15.02& 85.94\%& 63.58\%& 67.77\%& 66.20\%& \textbf{0.2215}& 29.4736 \\
 & InstanceControl (Hed)
 &   \textbf{0.8225}& \textbf{16.84}&  \textbf{92.11\%} &  \textbf{78.34\%}&  \textbf{85.45\%}& \textbf{85.09\%}&  0.1334 & \textbf{21.5091}\\
\bottomrule
\end{tabular}
\vspace{-8mm}
\label{tab:comparison_mutilin_coco}}
\end{table*}

\noindent\textbf{Qualitative Comparison.}
Qualitative comparisons with state-of-the-art methods are illustrated in \cref{fig:com_mutilin}.
While instance-labeled methods generally achieve satisfactory text-image consistency, layout-to-image frameworks like EliGen~\cite{zhang2025eligen} still struggle with complex instance overlaps (\eg, the case in the final row) due to the lack of visual conditions. Similarly, training-free controllable methods like DreamRenderer~\cite{zhou2025dreamrenderer} often fail to satisfy strict semantic constraints. In contrast, our method with instance labeling achieves superior text-image alignment.
For methods lacking instance labeling, such as FLUX ControlNet~\cite{flux2024}, we observe frequent attribute confusion in multi-instance scenarios. For example, in the second row, given the prompt: \texttt{"...a bright blue tang with a yellow tail... and further above it is a vibrant orange fish,"} FLUX ControlNet~\cite{flux2024} generates red and blue fish, directly contradicting the text description. Conversely, our method ensures precise attribute binding aligned with the text. These visual results highlight the superior capability of our model in fine-grained attribute control within complex multi-instance scenarios.

\begin{figure}[t!]
  \centering
  \includegraphics[width=0.99\linewidth]{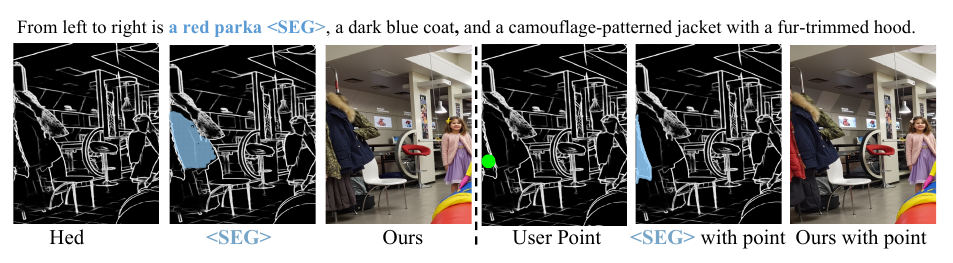}
  \vspace{-3mm}
  \caption{Visualization of the interactive mask correction.}
  \label{fig:mutilinstance_imc}
  \vspace{-1mm}
\end{figure}

\begin{table*}[t!]
\setlength{\tabcolsep}{5pt}
\centering
\caption{
Quantitative results with unified understanding and generation methods.
}
\vspace{-3mm}
\resizebox{0.97\textwidth}{!}{
\begin{tabular}{ccccccccc} 
\toprule

\multirow{3}{*}{Methods} & \multirow{3}{*}{MIoU$\uparrow$} &\multicolumn{5}{c}{Region-wise Quality} &\multicolumn{2}{c}{Global-wise Quality}\\
\cmidrule(lr){3-7}
\cmidrule(lr){8-9}
& & \makecell[c]{Local \\ CLIP$\uparrow$}  & Spatial$\uparrow$ & Color$\uparrow$ & Shape$\uparrow$  & Texture$\uparrow$  & IR$\uparrow$  &  FID$\downarrow$   \\
\midrule
Qwen-Image ControlNet~\cite{qwenimagecontol} &  0.8307 & 20.20 & 96.35\% & 89.58\% & 90.68\% & 91.13\% & 0.5143 & 116.1628 \\
Nano Banana~\cite{nano}  &0.8127  & 20.42& 97.06\% & 90.47\% & 91.02\% & 91.30\%& 0.5607 & 112.5264 \\
InstanceControl & \textbf{0.8834} & \textbf{20.54} & \textbf{97.25\%} &\textbf{92.21\%} & \textbf{92.12\%} & \textbf{92.49\%} &\textbf{0.5898}&  \textbf{100.8512} \\

\bottomrule
\end{tabular}
\label{tab:comparisoninstance_ugm}}
\vspace{-1mm}
\end{table*}

\noindent\textbf{Visualization of Learned Correspondences.}
Additionally, we visualize the learned correspondences and generated images in \cref{fig:com_instance_num}. We observe that our method accurately parses instance descriptions and predicts instance masks, facilitating controllable, fine-grained generation. More examples across varying instance numbers are provided in the supplementary material.

\noindent\textbf{Interactive Mask Correction.} 
We provide an interactive mechanism to rectify potential errors. When the refined mask $\mathbf{m}^{\mathit{rfn}}$ is suboptimal, users can provide points or bounding boxes to guide the mask prediction. As shown in \cref{fig:mutilinstance_imc}, such user guidance (\eg, \texttt{"a red parka"}) successfully corrects inaccurate masks, leading to higher-fidelity image generation.

\noindent\textbf{Relationship between Stage 1 and Stage 2.}
We evaluate Stage 1 using \textit{MIoU}, achieving a score of 0.71. To analyze its impact on Stage 2, we examine the correlation between Stage 1 \textit{MIoU} and Stage 2 \textit{Accuracy}, and observe a positive correlation; the scatter plot is provided in the supplementary material. Common Stage 1 failures include incomplete regions, localization offsets, and missed objects. Stage 2 is generally robust to incomplete regions, whereas more severe errors, such as localization offsets and missed objects, may cause incorrect generation. Our MRM mitigates these issues, improving \textit{Accuracy} from 87.97\% to 90.10\%, and further to 93.19\% with interactive correction.



\subsection{Ablation Study}

\noindent\textbf{Effect of Instance-level Association.}
As illustrated in \cref{sec:mutilinstance_association}, our InstanceControl establishes precise instance-level correspondences to achieve high-fidelity controllable generation. Compared with FLUX ControlNet~\cite{flux2024}, our method demonstrates significant improvements across the metrics in \cref{tab:comparison_mutilin_eval}, highlighting the effectiveness of our proposed instance-level association. Notably, even when utilizing only predicted masks without the mask refinement module, our framework consistently outperforms FLUX ControlNet~\cite{flux2024} (e.g., 87.97\% vs. 77.78\% average accuracy), further validating the impact of our association strategy.
We further conduct detailed ablation studies of the Shared SEG Token (SST).
%
%
%
We introduce the Shared SEG Token (SST) to ensure semantic consistency across multiple descriptions of the same instance, mitigating mask prediction errors during subsequent mentions. Quantitative results in \cref{tab:comparisoninstance_ief_SST} confirm that SST significantly enhances controllable image generation. We provide additional visualizations of its impact in the supplementary material.

\begin{table*}[t!]
\setlength{\tabcolsep}{5pt}
\centering
\caption{
Effect of the proposed Share SEG Token.}
\vspace{-3mm}
\resizebox{0.87\textwidth}{!}{
\begin{tabular}{ccccccccc} 
\toprule

 \multirow{3}{*}{SST} & \multirow{3}{*}{MIoU$\uparrow$} &\multicolumn{5}{c}{Region-wise Quality} &\multicolumn{2}{c}{Global-wise Quality}\\
\cmidrule(lr){3-7}
\cmidrule(lr){8-9}
 & & \makecell[c]{Local \\ CLIP$\uparrow$}  & Spatial$\uparrow$ & Color$\uparrow$ & Shape$\uparrow$  & Texture$\uparrow$  & IR$\uparrow$  &  FID$\downarrow$   \\
\midrule
$\times$&0.8214&18.31&93.16\% &85.89\% &86.46\% & 89.83\% & 0.2812&10.2983\\
$\checkmark$ & \textbf{0.8250} & \textbf{18.51}& \textbf{93.54\%} & \textbf{87.78\%} & \textbf{88.19\%} & \textbf{90.88\%} & \textbf{0.2847} & \textbf{10.0264} \\

\bottomrule
\end{tabular}
\vspace{-5mm}
\label{tab:comparisoninstance_ief_SST}}
\end{table*}

\begin{table*}[t!]
\centering
\caption{
Ablation study on the different masks for generation. Results demonstrate that the refined mask through the MRM, yields the optimal performance.
}
\vspace{-3mm}
\resizebox{1\textwidth}{!}{
\begin{tabular}{ccccccccc} 
\toprule

\multirow{3}{*}{Mask} & \multirow{3}{*}{MIoU$\uparrow$} &\multicolumn{5}{c}{Region-wise Quality
} &\multicolumn{2}{c}{Global-wise Quality}\\
\cmidrule(lr){3-7}
\cmidrule(lr){8-9}
& & \makecell[c]{Local \\ CLIP$\uparrow$}  & Spatial$\uparrow$ & Color$\uparrow$ & Shape$\uparrow$  & Texture$\uparrow$  & IR$\uparrow$  &  FID$\downarrow$   \\
\midrule
$\mathbf{m}^{\mathit{pred}}$   &0.8155 & 18.07 & 92.42\% & 84.88\% &85.48\% &89.10\% &0.2812 & 10.9284 \\
$s_i^{\mathit{pred}} \! \mathbf{m}_{i}^{\mathit{pred}} \! + \! (1 \! - \!s_i^{\mathit{pred}}) \mathbf{m}_i^{\mathit{att}}$ &0.8210 & 18.39 & 93.19\% & 85.91\% & 86.26\% & 89.99\% & 0.2833 & 10.0558
 \\
$\mathbf{m}^{\mathit{rfn}}$ &\textbf{0.8250} & \textbf{18.51}& \textbf{93.54\%} & \textbf{87.78\%} & \textbf{88.19\%} & \textbf{90.88\%} & \textbf{0.2847} & \textbf{10.0264} \\

\bottomrule
\end{tabular}
\label{tab:mutilinstance_mam}}
\vspace{-3mm}
\end{table*}

\noindent\textbf{Effect of Mask Refinement Module}.
In this subsection, we conduct ablation studies to evaluate the impact of the Mask Refinement Module (MRM). Specifically, we compare three mask variants: the raw predicted mask $\mathbf{m}^{\mathit{pred}}$, the fused mask $\mathbf{m}^{\mathit{fuse}}$, and our refined mask $\mathbf{m}^{\mathit{rfn}}$ generated by the MRM.
Direct reliance on $\mathbf{m}^{\mathit{pred}}$ inherently introduces prediction inaccuracies, which leads to degraded performance. To mitigate this, we consider a baseline fusion strategy, $\mathbf{m}^{\mathit{fuse}}$, defined by a score-based interpolation: $\mathbf{m}_{i}^{\mathit{fuse}} = s_i^{\mathit{pred}}  \mathbf{m}_{i}^{\mathit{pred}} + (1 - s_i^{\mathit{pred}}) \mathbf{m}_i^{\mathit{att}}$. Although $\mathbf{m}^{\mathit{fuse}}$ offers some improvements as shown in \cref{tab:mutilinstance_mam}, it lacks the robustness for complex scenarios. In contrast, our MRM yields a superior refined mask by adaptively integrating multiple cues, including the predicted mask $\mathbf{m}^{\mathit{pred}}_i$, the attention-based mask $\mathbf{m}^{\mathit{att}}_{i}$, the confidence score $s_i^{\mathit{pred}}$, and the image latent features $\mathbf{H}_{\mathit{img}}$. The consistent performance gains reported in \cref{tab:mutilinstance_mam} validate the effectiveness of our refinement mechanism.


\section{Conclusion}

In this paper, we introduce InstanceControl, a novel framework for multi-instance controllable generation, specifically designed to mitigate attribute confusion in complex scenes. Unlike previous approaches that rely on labor-intensive manual instance labeling, our method leverages a Vision-Language Model (VLM) to automatically establish correspondences between text prompts and visual conditions. Recognizing that initial predicted masks may contain noise, we introduce a mask refinement strategy that dynamically adjusts masks during the generation process to achieve more accurate generation. Extensive experiments across multiple benchmarks and visual conditions demonstrate that our approach achieves superior fidelity and precise instance-level control. By significantly outperforming current state-of-the-art methods in multi-instance scenarios, our framework proves highly effective even without the need for instance labeling.


\section*{Acknowledgements}
This work was supported by the National Cyber Security-National Science and Technology Major Project under Grant No. 2026ZD1500500 and the National Natural Science Foundation of China (NSFC) under Grant No. 62501191.

%
%
\bibliographystyle{splncs04}
\bibliography{main}
\end{document}